\documentclass[letterpaper]{article} 
\usepackage{aaai2026}  
\usepackage{amsfonts}
\usepackage{times}  
\usepackage{helvet}  
\usepackage{courier}  
\usepackage[hyphens]{url}  
\usepackage{graphicx} 
\usepackage{natbib}  
\usepackage{caption} 
\usepackage{algorithm}
\usepackage{algorithmic}
\usepackage{amsmath}
\usepackage{multirow}
\usepackage{graphicx}
\usepackage{diagbox}
\usepackage{booktabs}
\usepackage{pdfpages}
\usepackage{subcaption}

\usepackage{newfloat}
\usepackage{listings}
\DeclareCaptionStyle{ruled}{labelfont=normalfont,labelsep=colon,strut=off} 
\floatstyle{ruled}
\newfloat{listing}{tb}{lst}{}
\floatname{listing}{Listing}
\title{A Multi-dimensional Semantic Surprise Framework Based on Low-Entropy Semantic Manifolds for Fine-Grained Out-of-Distribution Detection}

\author {
    Ningkang Peng,
    Yuzhe Mao,
    Yuhao Zhang,
    Linjin Qian,
    Qianfeng Yu,
    Yanhui Gu\thanks{Corresponding author.},
    Yi Chen\footnotemark[1],
    Li Kong\footnotemark[1]
}
\affiliations {
    School of Computer and Electronic Information, Nanjing Normal University, China\\
    gu@njnu.edu.cn, 
    cs\_chenyi@njnu.edu.cn, 
    kongli@nnu.edu.cn
}

\begin{document}
\nocopyright
\maketitle

\begin{abstract}
Out-of-Distribution (OOD) detection is a cornerstone for the safe deployment of AI systems in the open world. However, existing methods treat OOD detection as a binary classification problem, a cognitive flattening that fails to distinguish between semantically close (Near-OOD) and distant (Far-OOD) unknown risks. This limitation poses a significant safety bottleneck in applications requiring fine-grained risk stratification. To address this, we propose a paradigm shift from a conventional probabilistic view to a principled information-theoretic framework. We formalize the core task as quantifying the Semantic Surprise of a new sample and introduce a novel ternary classification challenge: In-Distribution (ID) vs. Near-OOD vs. Far-OOD. The theoretical foundation of our work is the concept of Low-Entropy Semantic Manifolds, which are explicitly structured to reflect the data's intrinsic semantic hierarchy. To construct these manifolds, we design a Hierarchical Prototypical Network. We then introduce the Semantic Surprise Vector (SSV), a universal probe that decomposes a sample's total surprise into three complementary and interpretable dimensions: conformity, novelty, and ambiguity. To evaluate performance on this new task, we propose the Normalized Semantic Risk (nSR), a cost-sensitive metric. Experiments demonstrate that our framework not only establishes a new state-of-the-art (sota) on the challenging ternary task, but its robust representations also achieve top results on conventional binary benchmarks, reducing the False Positive Rate by over 60\% on datasets like LSUN.
\end{abstract}
\section{Introduction}

Out-of-Distribution (OOD) detection is a prerequisite for the safe deployment of machine learning models in open-world applications, from autonomous driving to intelligent medical care \citep{c:48,c:49}. The fundamental goal is to identify novel inputs that deviate from the training distribution, thereby preventing catastrophic mispredictions \citep{c:50}. The research community has made significant progress, particularly with representation-learning-based methods that learn a compact embedding space to separate in-distribution (ID) samples from unknowns \citep{c:41, c:40}. State-of-the-art approaches like PALM \citep{c:35} can even learn fine-grained, multi-prototype representations for known classes, enhancing their ability to reject unseen inputs \citep{c:26, c:34}.

\begin{figure*}[htbp]
    \centering 

    \begin{subfigure}[b]{1\columnwidth}
        \centering
        \includegraphics[width=\textwidth]{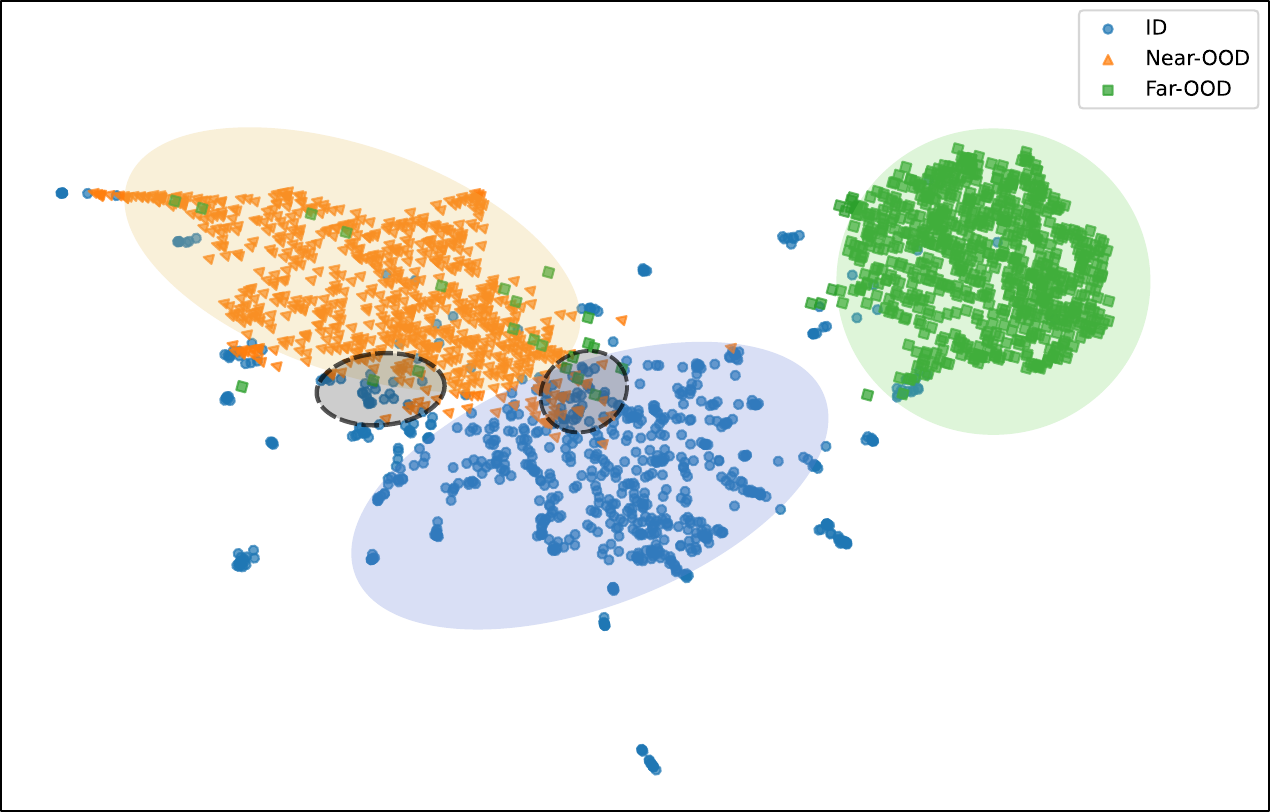} 
        \caption{PALM}
        \label{fig:vis_palm}
    \end{subfigure}
    \hfill 
    \begin{subfigure}[b]{1\columnwidth}
        \centering
        \includegraphics[width=\textwidth]{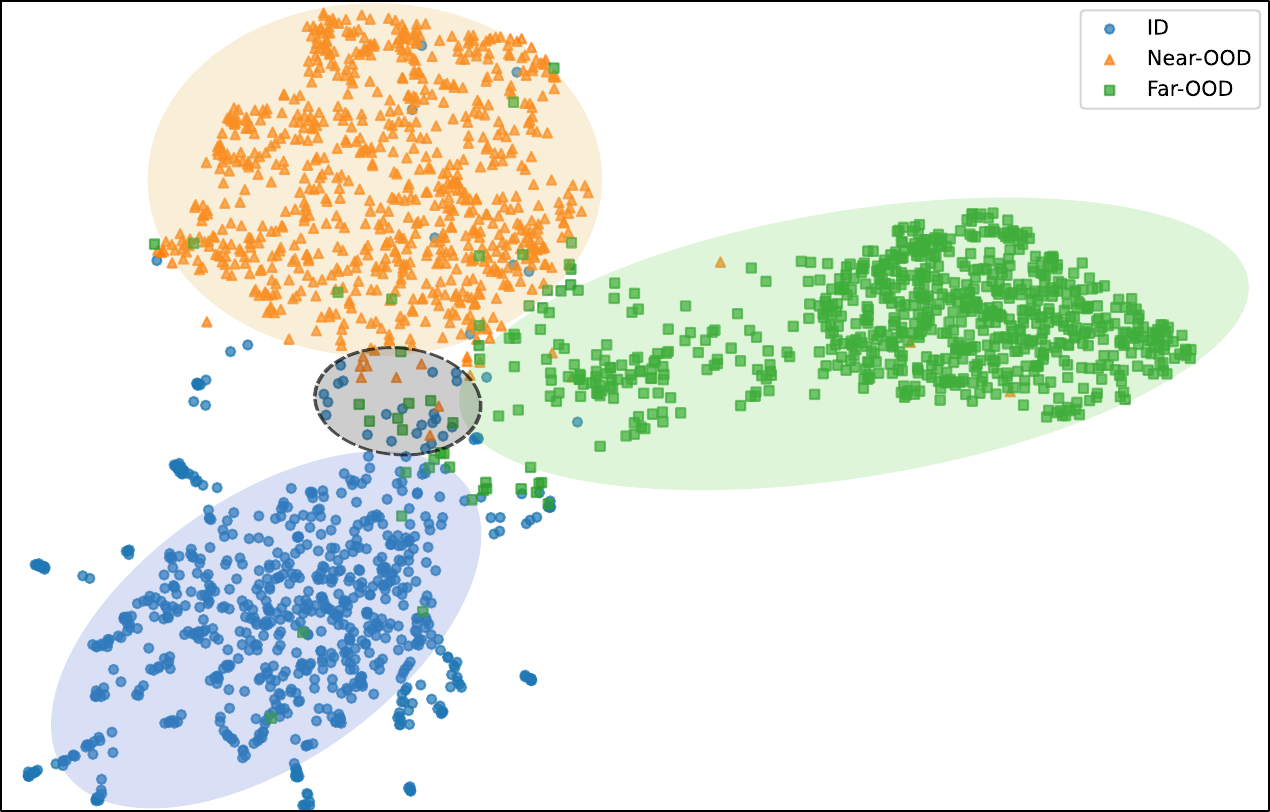}
        \caption{Our}
        \label{fig:vis_ours}
    \end{subfigure}

    \caption{Visualization of feature spaces. (a) The PALM suffers from cognitive flattening, evidenced by the severe overlap between ID (blue) and Near-OOD (orange) clusters (highlighted by black ellipses). (b) Our method creates a significantly better-separated manifold, crucial for distinguishing fine-grained OOD risks.}
    \label{fig:manifold_visualization}
\end{figure*}

However, a profound limitation underlies these achievements: existing OOD frameworks universally treat OOD detection as a binary classification problem. This simplification leads to a cognitive flattening, where the model is incapable of distinguishing between semantically close unknowns (e.g., a tiger for a model trained on CIFAR-100 animals, which we term \textit{Near-OOD}) and semantically distant ones (e.g., a keyboard, or \textit{Far-OOD}). As visualized in Figure~\ref{fig:manifold_visualization}, this forces a severe overlap between the feature distributions of in-distribution and Near-OOD samples, where both are crudely collapsed into a single OOD category \citep{c:25}. In safety-critical systems that demand fine-grained risk stratification, this inability to assess the nature of an unknown constitutes a major safety bottleneck \citep{c:40, c:47}.

We posit that the root of this deficiency lies in a failure to manage Information Entropy within the representation space. While existing methods excel at sculpting low-entropy regions for known ID classes (i.e., compact clusters), they treat the vast, heterogeneous world of the unknown as a single, undifferentiated high-entropy region \citep[e.g.,][]{c:53,c:54,c:35}. By forcibly mapping all OOD samples into this chaotic space, regardless of their semantics, the model is deprived of the structural information needed for further judgment \citep[cf.][]{c:55}. This uncontrolled entropy makes cognitive flattening an inevitable outcome.
To resolve this dilemma, we advocate for a paradigm shift from probabilistic judgment to an information-theoretic framework centered on quantifying Semantic Surprise. The core idea is to actively reduce entropy in the unknown space by imposing a meaningful semantic structure. Our approach unfolds in three stages. First, we redefine the problem as a ternary classification task: ID vs. Near-OOD vs. Far-OOD. Second, to provide a geometric foundation for this task, we introduce the concept of Low-Entropy Semantic Manifolds and propose a Hierarchical Prototypical Network to construct them. This network is trained with a novel objective that organizes subclass prototypes according to their shared superclass semantics. Finally, to probe these structured manifolds, we develop the Semantic Surprise Vector (SSV), which decomposes a sample's surprise into three interpretable dimensions: conformity, novelty, and ambiguity.

Our main contributions are as follows:
\begin{enumerate}
    \item \textbf{Theoretical Paradigm Shift}: We are the first to identify cognitive flattening as a key bottleneck and reframe OOD detection as an information-theoretic problem of quantifying Semantic Surprise, leading to a new ID vs. Near-OOD vs. Far-OOD ternary classification challenge.
    \item \textbf{Methodological Framework}: We propose a complete framework comprising a Hierarchical Prototypical Network to sculpt Low-Entropy Semantic Manifolds and a Semantic Surprise Vector to perform multi-dimensional, interpretable risk diagnosis.
    \item \textbf{Comprehensive Evaluation}: We introduce a new cost-sensitive metric, the Normalized Semantic Risk, for this ternary task and conduct extensive experiments that establish a new sota, demonstrating our framework's superior performance and robustness.
\end{enumerate}

\section{Related Work}
\label{sec:related_work}

The field of Out-of-Distribution detection has evolved from post-hoc scoring of pre-trained models to proactively shaping the feature space geometry. A comprehensive review of this trajectory is provided in Appendix A. We argue that the success of modern geometric methods can be understood as an implicit pursuit of Low-Entropy Manifolds for in-distribution (ID) classes. For instance, methods enforcing hyperspherical uniformity \citep{c:31,c:34,zou2025provable} or using contrastive objectives are essentially attempting to minimize the geometric volume and thus the entropy of ID representations.

However, this pursuit has been incomplete. Critically, existing methods focus exclusively on structuring the ID space, treating the vast OOD space as a single, high-entropy void. Furthermore, they lack a mechanism to encode semantic relationships; for example, the manifold for 'truck' is not explicitly encouraged to be closer to 'car' than to 'cat'. This lack of semantic hierarchy is the fundamental reason they fail to distinguish Near-OOD from Far-OOD samples, leading to the cognitive flattening bottleneck.
To transcend this, our work represents the first synthesis of the Information Bottleneck (IB) principle \citep{tishby2000information} with hierarchical learning to explicitly construct a Low-Entropy Semantic Manifold that structures both the known and the unknown. While prior works have used hierarchies for classification \citep{c:62, c:63,wallin2025prohoc}, they do not address the core OOD challenge of structuring the unknown space for fine-grained risk stratification.

\section{Methodology}
\begin{figure*}[htbp]
    \centering
    \includegraphics[width=0.9\textwidth]{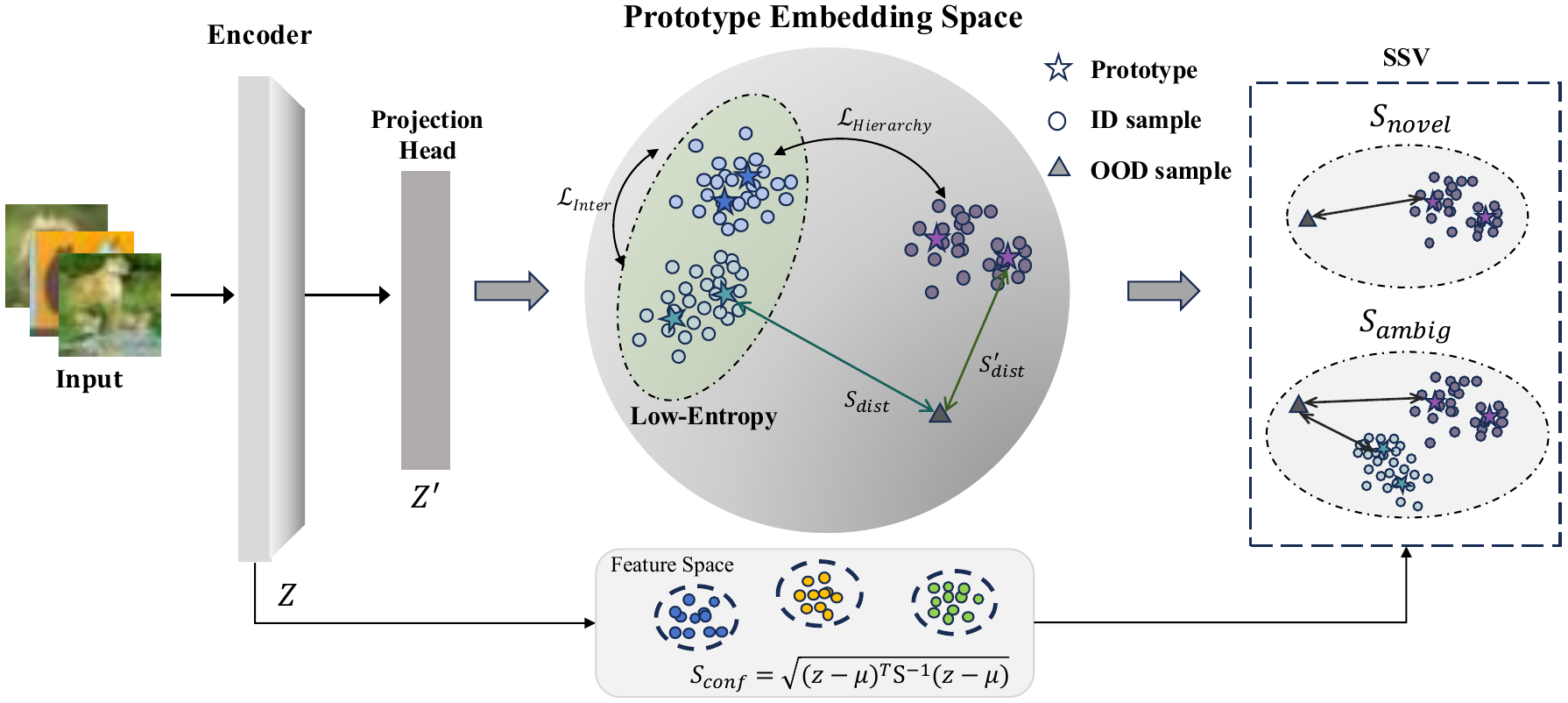}
    \caption{Overview of our proposed framework. We learn a hierarchically-structured manifold on a hypersphere and probe it using the multi-dimensional SSV to inform a final risk classifier.}
\label{fig:framework_overview}
\end{figure*}

Our methodology for fine-grained OOD detection is built upon a novel information-theoretic framework. We first introduce the theoretical foundations of our approach, followed by the specific mechanisms for manifold shaping, probing, and evaluation. An overview of the entire framework is illustrated in Figure~\ref{fig:framework_overview}.

\subsection{Theoretical Foundation: The Low-Entropy Semantic Manifold}
\label{sec:theoretical_framework}
To transcend cognitive flattening, we posit that the solution lies in shaping a feature space with an ordered, predictable structure, which we term a Low-Entropy Semantic Manifold. Its geometric topology must intuitively reflect the data's intrinsic semantic hierarchy. We ground its construction in a Hierarchical Information Bottleneck (HIB) framework.
The classic IB principle seeks to learn a representation $\mathbf{z}$ of an input $x$ that is maximally compressed yet maximally informative about a target $y$. This is typically formulated as minimizing the Lagrangian:
\begin{equation}
\mathcal{L}_{\text{IB}} = I(\mathbf{Z}; X) - \beta I(\mathbf{Z}; Y),
\end{equation}
where the first term enforces compression and the second term preserves task-relevant information.

In our setting, the label possesses a two-level hierarchy: $y = (M, c)$. By decomposing the information term $I(\mathbf{Z}; Y)$ using the chain rule, we extend the classic IB objective to our Hierarchical Information Bottleneck (HIB) objective. We formulate it as a loss function to be minimized:
\begin{equation}
\mathcal{L}_{\text{HIB}} = \underbrace{I(\mathbf{Z}; X)}_{\text{Compression Term}} - \beta_M \underbrace{I(\mathbf{Z}; M)}_{\text{Preservation Term 1}} - \beta_c \underbrace{I(\mathbf{Z}; c|M)}_{\text{Preservation Term 2}}.
\label{eq:hib}
\end{equation}
Minimizing this loss function involves a trade-off: the first term, $I(\mathbf{Z};X)$, is minimized to compress irrelevant details from the input, while the negative mutual information terms, $-I(\mathbf{Z};M)$ and $-I(\mathbf{Z};c|M)$, are also minimized, which is equivalent to maximizing the preservation of task-relevant hierarchical information. As direct optimization is difficult, we geometrize this objective into three design principles.

\paragraph{Principle 1: Subclass Compactness.} To maximize the microscopic information $I(\mathbf{Z}; c|M)$, embeddings from the same subclass $c$ must form a tight, low-entropy cluster. Formally, the average distance between any two embeddings drawn from the same subclass should be minimal:
\begin{equation}
    \mathbb{E}_{\mathbf{z}_i, \mathbf{z}_j \sim c} [d(\mathbf{z}_i, \mathbf{z}_j)] \le \epsilon_c.
\end{equation}

\paragraph{Principle 2: Superclass Cohesion.} To maximize the macroscopic information $I(\mathbf{Z}; M)$, manifolds of different subclasses ($c_i, c_j$) belonging to the \textit{same superclass} $M$ must be geometrically adjacent. This encourages intra-superclass consistency:
\begin{equation}
    \epsilon_c < \mathbb{E}_{\mathbf{z}_i \sim c_i, \mathbf{z}_j \sim c_j; c_i, c_j \subset M, i \neq j} [d(\mathbf{z}_i, \mathbf{z}_j)] \le \epsilon_M.
\end{equation}

\paragraph{Principle 3: Superclass Separation.} To further maximize $I(\mathbf{Z}; M)$ by ensuring discriminability between superclasses, the manifolds belonging to different superclasses ($M_1, M_2$) must be separated by a significant, low-entropy gap. Cohesion and Separation are thus two complementary geometric facets of maximizing macroscopic information:
\begin{equation}
    \mathbb{E}_{\mathbf{z}_i \in M_1, \mathbf{z}_j \in M_2; M_1 \neq M_2} [d(\mathbf{z}_i, \mathbf{z}_j)] > \epsilon_M + \Delta.
\end{equation}

\paragraph{Manifold Quality Metrics.}
To quantitatively diagnose the geometric health of any feature space, we define three metrics: compactness, cohesion, and separation, which correspond directly to our three design principles (see Appendix for formal definitions).

\subsection{Manifold Shaping: The Hierarchical Prototypical Network}
\label{sec:manifold_shaping}
Building upon recent advances in multi-prototype learning, particularly inspired by frameworks like PALM\cite{c:35}, we design a Hierarchical Prototypical Network to realize these geometric principles. The model comprising a backbone $f_{\theta}$ and a projection head $g_{\phi}$, maps an input $x$ to a normalized embedding $\mathbf{z} = g_{\phi}(f_{\theta}(x))$. We sculpt the manifold using a composite loss function operating at two levels: the sample level and the prototype level.

\paragraph{Sample-Level Compactness.}
To enforce subclass compactness while capturing intra-class diversity, we model each subclass $c$ as a mixture of $K$ learnable, L2-normalized prototypes $\{\mathbf{p}_k^c\}_{k=1}^K$. The soft assignment weights $w_{i,k}^c$ for a sample $\mathbf{z}_i$ to these prototypes are computed via an online Sinkhorn-Knopp algorithm \citep{cuturi2013sinkhorn}, where $\epsilon$ is a regularization parameter.
Our training objective is to perform Maximum Likelihood Estimation (MLE) on this class-conditional mixture model. Specifically, we aim to maximize the posterior probability $p(y_i=c(i)|\mathbf{z}_i)$ (equivalent to minimizing its negative log-likelihood (NLL)). Using Bayes' theorem, the posterior is given by:
\begin{equation}
p(y_i=c | \mathbf{z}_i) = \frac{p(\mathbf{z}_i | y_i=c) p(y_i=c)}{\sum_{j=1}^C p(\mathbf{z}_i | y_i=j) p(y_i=j)},
\end{equation}
where $p(\mathbf{z}_i | y_i=c) = \sum_{k=1}^K w_{i,k}^c p(\mathbf{z}_i | \mathbf{p}_k^c)$ is our mixture model likelihood, and $p(\mathbf{z}_i | \mathbf{p}_k^c) \propto \exp(\mathbf{z}_i^\top \mathbf{p}_k^c / \tau)$. By taking the negative log of this posterior and assuming a uniform prior over classes, we arrive at the final loss function, which we term $\mathcal{L}_{\text{MLE}}$:
\begin{equation}
\mathcal{L}_{\text{MLE}} = - \frac{1}{N} \sum_{i=1}^{N} \log \frac{\sum_{k=1}^K w_{i,k}^{c(i)} \exp(\mathbf{z}_i^\top \mathbf{p}_k^{c(i)} / \tau)}{\sum_{j=1}^C \sum_{k'=1}^K w_{i,k'}^{j} \exp(\mathbf{z}_i^\top \mathbf{p}_{k'}^j / \tau)},
\label{eq:loss_mle}
\end{equation}
where $c(i)$ is the ground-truth subclass for $\mathbf{z}_i$, and $\tau$ is a temperature parameter.

\paragraph{Prototype-Level Structure.}
We introduce two contrastive losses on the prototypes themselves. First, an Inter-Prototype Contrastive Loss ($\mathcal{L}_{\text{Inter-Proto}}$) reinforces Principle 1 and part of Principle 3. For an anchor prototype $\mathbf{p}_k^c$, its positive set consists of other prototypes from the same subclass $\{\mathbf{p}_{k'}^c\}_{k' \neq k}$, and its negative set contains all prototypes from all other subclasses. This takes the standard InfoNCE form:
\begin{align}
&\mathcal{L}_{\text{Inter-Proto}} = {}  - \frac{1}{CK} \sum_{c=1}^C \sum_{k=1}^K \frac{1}{K-1} \sum_{k' \neq k} \notag \\
& \quad \log \frac{\exp(\mathbf{p}_k^{c\top} \mathbf{p}_{k'}^c / \tau_p)}{\sum_{(j,k'') \neq (c,k)} \exp(\mathbf{p}_k^{c\top} \mathbf{p}_{k''}^j / \tau_p)}.
\label{eq:loss_proto_contra}
\end{align}
Second, our core innovation is a hierarchical prototype loss ($\mathcal{L}_{Hierarchy}$), which explicitly enforces Principle 2 and refines Principle 3. For an anchor prototype $\mathbf{p}_k^c$, its positive set $\mathcal{P}(c,k)$ contains all prototypes from different subclasses within the same superclass, and its negative set $\mathcal{N}(c,k)$ contains all prototypes from other superclasses. The loss is formulated as:
\begin{align}
& \mathcal{L}_{\text{Hierarchy}} = {}  - \frac{1}{CK} \sum_{c,k} \frac{1}{|\mathcal{P}(c,k)|} \sum_{\mathbf{p}_p \in \mathcal{P}(c,k)} \notag \\
& \quad \log \frac{\exp(\mathbf{p}_k^{c\top} \mathbf{p}_p / \tau_h)}{\exp(\mathbf{p}_k^{c\top} \mathbf{p}_p / \tau_h) + \sum\limits_{\mathbf{p}_n \in \mathcal{N}(c,k)} \exp(\mathbf{p}_k^{c\top} \mathbf{p}_n / \tau_h)}.
\label{eq:loss_hierarchy}
\end{align}
This loss creates attractive forces between sibling subclass manifolds (cohesion) and repulsive forces between manifolds of different superclasses (separation). The final objective is 
\begin{equation}
\mathcal{L}_{\text{Total}} = \mathcal{L}_{\text{MLE}} + \lambda_1 \mathcal{L}_{\text{Inter-Proto}} + \lambda_2 \mathcal{L}_{\text{Hierarchy}}.
\end{equation}
Through this composite loss, the network parameters $(\theta, \phi)$ are optimized to resolve a controlled tension: $\mathcal{L}_{\text{Inter-Proto}}$ exerts a gradient signal that pushes embeddings away from sibling-class prototypes to ensure subclass purity, while $\mathcal{L}_{\text{Hierarchy}}$ exerts an opposing signal that pulls them closer to create superclass cohesion. The balance of these opposing gradient signals on the sample embeddings is what ultimately engineers the desired hierarchical manifold. In our experiments, we found that a simple weighting of $\lambda_1=\lambda_2=1$ provides a robust balance, though we acknowledge that tuning these values could further refine the manifold structure.

\paragraph{Prototype Updating.}
Crucially, the prototypes themselves are not trainable parameters but are updated via an Exponential Moving Average (EMA)\cite{ema} of sample embeddings. The gradients from the prototype-level losses update the network parameters. The full mechanism is detailed in Appendix. The update rule for a prototype $\mathbf{p}_k^c$ is:
\begin{equation}
\mathbf{p}_k^c \leftarrow \text{Normalize}\left(\alpha \mathbf{p}_k^c + (1-\alpha) \sum_{i=1}^{B} \mathbb{I}(y_i=c) w_{i,k}^c \mathbf{z}_i\right).
\label{eq:ema_update}
\end{equation}

\subsection{Manifold Probing: The Semantic Surprise Vector}
\label{sec:ssv}

After shaping the manifold, we propose a diagnostic probe SSV to assess the risk of a new sample, $\mathbf{z}_{\text{new}}$. The SSV deconstructs the total surprise into three complementary components, each providing a distinct geometric interpretation of risk (see Appendix for the probabilistic motivation). To ensure a universal diagnostic framework, all SSV computations use Euclidean distance. The only adaptation is the choice of concept representatives: for our method, we use the learned prototypes; for baselines, we use their class centroids.

\paragraph{1. Conformity Surprise ($S_{\text{conf}}$).}
This component measures how much a sample deviates from the global ID data statistics. We model the global feature distribution as a multivariate Gaussian and define $S_{\text{conf}}$ using the Mahalanobis distance, a natural metric for such models:
\begin{equation}
S_{\text{conf}}(\mathbf{z}_{\text{new}}) = \sqrt{(\mathbf{z}_{\text{new}} - \boldsymbol{\mu}_{\text{global}})^\top \mathbf{\Sigma}_{\text{global}}^{-1} (\mathbf{z}_{\text{new}} - \boldsymbol{\mu}_{\text{global}})},
\label{eq:s_conf}
\end{equation}
where $\boldsymbol{\mu}_{\text{global}}$ and $\mathbf{\Sigma}_{\text{global}}$ are the mean and regularized covariance of the ID training features.

\paragraph{2. Novelty Surprise ($S_{\text{novel}}$).}
This component quantifies if a sample falls into a knowledge gap far from any known concept. We define it as the Euclidean distance to the nearest concept representative, $\mathbf{r}$, from the set of all representatives $\mathcal{R}$:
\begin{equation}
S_{\text{novel}}(\mathbf{z}_{\text{new}}) = \min_{\mathbf{r} \in \mathcal{R}} \|\mathbf{z}_{\text{new}} - \mathbf{r}\|_2.
\label{eq:s_novel}
\end{equation}

\paragraph{3. Ambiguity Surprise ($S_{\text{ambig}}$).}
This component captures the model's indecision when a sample is equidistant from multiple distinct concepts. It is defined as the ratio of the distances to the nearest representative ($\mathbf{r}_1$) and the second-nearest representative from a different class ($\mathbf{r}_2$):
\begin{equation}
S_{\text{ambig}}(\mathbf{z}_{\text{new}}) = \frac{\min_{\mathbf{r} \in \mathcal{R}} \|\mathbf{z}_{\text{new}} - \mathbf{r}\|_2}{\min_{\mathbf{r} \in \mathcal{R} \setminus \mathcal{R}_{c_1}} \|\mathbf{z}_{\text{new}} - \mathbf{r}\|_2},
\label{eq:s_ambig}
\end{equation}
where $\mathcal{R}_{c_1}$ is the set of representatives for the subclass of $\mathbf{r}_1$. As this ratio approaches 1, ambiguity is maximal. This deconstruction provides a rich, multi-faceted diagnostic report.

\subsection{Risk-Aware Evaluation: The Normalized Semantic Risk}
\label{sec:nsr}
Traditional metrics like accuracy are inadequate for our task, as they treat all errors equally. We propose the Normalized Semantic Risk, a metric grounded in Bayesian Decision Theory that quantifies risk by using a cost matrix derived from three rational risk principles (e.g., \textit{Security Boundary Precedence}). The nSR normalizes a model's total empirical risk by that of a naive baseline model that always predicts ID. This yields a final score where lower is better, indicating superior risk-aware decision-making. The full derivation, principles, and cost matrix are detailed in Appendix.
The nSR for a model's predictions on a test set is calculated as:
\begin{equation}
\text{nSR} = \frac{R_{\text{total}}}{R_{\text{max}}} = \frac{\sum_{i=1}^{N} C(y_i^{\text{true}}, y_i^{\text{pred}})}{5 N_{N} + 6 N_{F}}.
\label{eq:nsr_final_appendix}
\end{equation}
where $N_{N}$ and $N_{F}$ are the number of Near-OOD and Far-OOD samples, respectively, and $C$ is the cost function.

\section{Experiments}
\label{sec:experiments}

\subsection{Experimental Setup}
\label{sec:exp_setup}
We conduct our core experiments using CIFAR-100 as the ID dataset, with CIFAR-10 as Near-OOD and SVHN as Far-OOD. Our framework utilizes a ResNet-34 backbone to produce feature embeddings. These embeddings are then processed by our proposed method to generate a three-dimensional SSV. To obtain the final trinary risk classification, the SSV is used as input to a LightGBM\cite{lightgbm} classifier. We evaluate performance using our proposed nSR metric and the Macro F1-score\cite{macrof1}. A comprehensive description of all datasets, implementation details, baseline configurations, and a full list of hyperparameters is provided in Appendix.

\begin{table*}[htbp]
    \centering
    \renewcommand{\arraystretch}{0.95}
    \begin{tabular}{llccccc}
        \toprule
        \textbf{Backbone} & \textbf{Classifier} & \textbf{nSR} $\downarrow$ & \textbf{F1 (ID)} $\uparrow$ & \textbf{F1 (Near)} $\uparrow$ & \textbf{F1 (Far)} $\uparrow$ & \textbf{Macro F1} $\uparrow$ \\
        \midrule
        \multirow{3}{*}{CSI} & K-Means & 0.6117 & 0.21 & 0.49 & 0.14 & 0.28 \\
        & Oracle & 0.5589 & - & - & - & 0.29 \\
        & SSV & 0.4384 & 0.17 & 0.38 & \underline{0.49} & 0.35 \\
        \midrule
        \multirow{3}{*}{SSD+} & K-Means & 0.7454 & 0.30 & 0.31 & 0.19 & 0.27 \\
        & Oracle & 0.2943 & - & - & - & 0.43 \\
        & SSV & \underline{0.3635} & 0.40 & 0.34 & \textbf{0.62} & \underline{0.45} \\
        \midrule
        \multirow{3}{*}{KNN+} & K-Means & 0.6700 & 0.25 & 0.45 & 0.16 & 0.29 \\
        & Oracle & 0.4392 & - & - & - & 0.36 \\
        & SSV & 0.5196 & 0.30 & \underline{0.56} & 0.35 & 0.40 \\
        \midrule
        \multirow{3}{*}{CIDER} & K-Means & 0.8311 & 0.31 & 0.20 & 0.15 & 0.22 \\
        & Oracle & 0.2789 & - & - & - & 0.57 \\
        & SSV & 0.4370 & 0.47 & 0.46 & 0.33 & 0.42 \\
        \midrule
        \multirow{3}{*}{NPOS} & K-Means & 0.7326 & 0.37 & 0.37 & 0.00 & 0.25 \\
        & Oracle & 0.3075 & - & - & - & 0.53 \\
        & SSV & 0.4294 & \underline{0.48} & 0.44 & 0.34 & 0.42 \\
        \midrule
        \multirow{3}{*}{PALM} & K-Means & 0.5887 & 0.43 & 0.44 & 0.27 & 0.38 \\
        & Oracle & 0.3109 & - & - & - & 0.54 \\
        & SSV & 0.4265 & \underline{0.48} & 0.52 & 0.30 & 0.43 \\
        \midrule
        \multirow{3}{*}{\textbf{Ours}} & K-Means & 0.7261 & 0.34 & 0.38 & 0.18 & 0.30 \\
        & Oracle & 0.2881 & - & - & - & 0.58 \\
        & SSV & \textbf{0.3268} & \textbf{0.50} & \textbf{0.64} & 0.45 & \textbf{0.53} \\
        \bottomrule
    \end{tabular}
    \caption{Comprehensive performance comparison on the trinary risk stratification task. Our full framework achieves the best Macro F1 score among all deployable methods. Oracle methods represent theoretical upper bounds and are excluded from best/second-best rankings. Best result in each column is in \textbf{bold}, second best is \underline{underlined}.}
    \label{tab:main_results_final}
\end{table*}

\subsection{Core Trinary Task Performance}
In our experiments, we benchmark our method against a comprehensive suite of sota baselines, including: MSP \citep{c:25}, Vim \citep{wang2022vim}, ODIN \citep{c:43}, Energy \citep{c:37}, VOS \citep{c:29}, CSI \citep{c:40}, SSD+ \citep{c:42}, KNN+ \citep{c:51}, CIDER \citep{c:34}, NPOS \citep{c:30}, and PALM \citep{c:35}. Our analysis proceeds in two stages: first, an in-depth geometric diagnosis of a representative subset of these methods, followed by the full performance comparison on the downstream trinary risk stratification task.

\paragraph{Geometric Diagnosis: Diagnosing and Preventing Semantic Collapse.}
Our central hypothesis that a well-structured manifold is a prerequisite for fine-grained OOD detection is empirically validated by diagnosing the geometric health of the learned feature spaces  in Table~\ref{tab:manifold_quality} and Table~\ref{tab:prototype_angles}. The results reveal a semantic collapse in baseline methods, a flaw our approach successfully prevents. Table~\ref{tab:manifold_quality} shows a critical trade-off in existing methods: baselines like PALM achieve high Separation scores but at the expense of Superclass Cohesion. Their significantly worse Cohesion scores provide quantitative proof of their failure to group semantically related classes. In stark contrast, our method achieves the best Cohesion and Compactness, reflecting a more sophisticated geometric arrangement that prioritizes meaningful, hierarchical organization over simple, maximal dispersion. This structural superiority is further explained at the prototype level in Table~\ref{tab:prototype_angles}. While for PALM, the near-identical intra-superclass ($68.64^\circ$) and inter-superclass ($71.07^\circ$) angles are symptomatic of a flat, non-hierarchical manifold, our method performs a powerful global compression. Crucially, this compression preserves the correct semantic ordering, with the intra-superclass angle ($48.86^\circ$) remaining smaller than the inter-superclass angle ($49.85^\circ$). This combined evidence from both the manifold and prototype levels demonstrates that our framework engineers a fundamentally more structured and lower-entropy organization.

\begin{table}[htbp]
    \centering
    \renewcommand{\arraystretch}{0.8}
    \begin{tabular}{l c c c}
        \toprule
        Method & Compactness $\downarrow$ & Cohesion $\downarrow$ & Separation $\uparrow$ \\
        \midrule
        SSD+ & 0.8071 & 1.0247 & 1.0207 \\
        KNN+ & 0.7269 & 1.0464 & \underline{1.0455} \\
        CIDER & 0.6263 & \underline{0.8812} & 0.8805 \\
        NPOS & 0.5895 & 0.8974 & 0.8958 \\
        PALM & \underline{0.4536} & 1.1670 & \textbf{1.1684} \\
        Ours & \textbf{0.3595} & \textbf{0.8650} & 0.8632 \\
        \bottomrule
    \end{tabular}
    \caption{Quantitative comparison of Manifold Quality Metrics. Our method excels in forming compact and cohesive semantic structures, while baselines sacrifice cohesion for raw separation.}
    \label{tab:manifold_quality}
\end{table}

\begin{table}[htbp]
    \centering
    \renewcommand{\arraystretch}{1.2}
    \begin{tabular}{l c c}
        \toprule
        Method & Avg. Intra Angle $\downarrow$ & Avg. Inter Angle $\uparrow$ \\
        \midrule
        PALM & 68.64$^\circ$ & 71.07$^\circ$ \\
        Ours & \textbf{48.86$^\circ$} & \textbf{49.85$^\circ$} \\
        \bottomrule
    \end{tabular}
    \caption{Analysis of average inter-prototype angles (in degrees). Our method demonstrates powerful global compression while preserving the correct semantic angular ordering (Intra $<$ Inter).}
    \label{tab:prototype_angles}
\end{table}

\begin{figure}[htbp]
    \centering
    \includegraphics[width=\columnwidth]{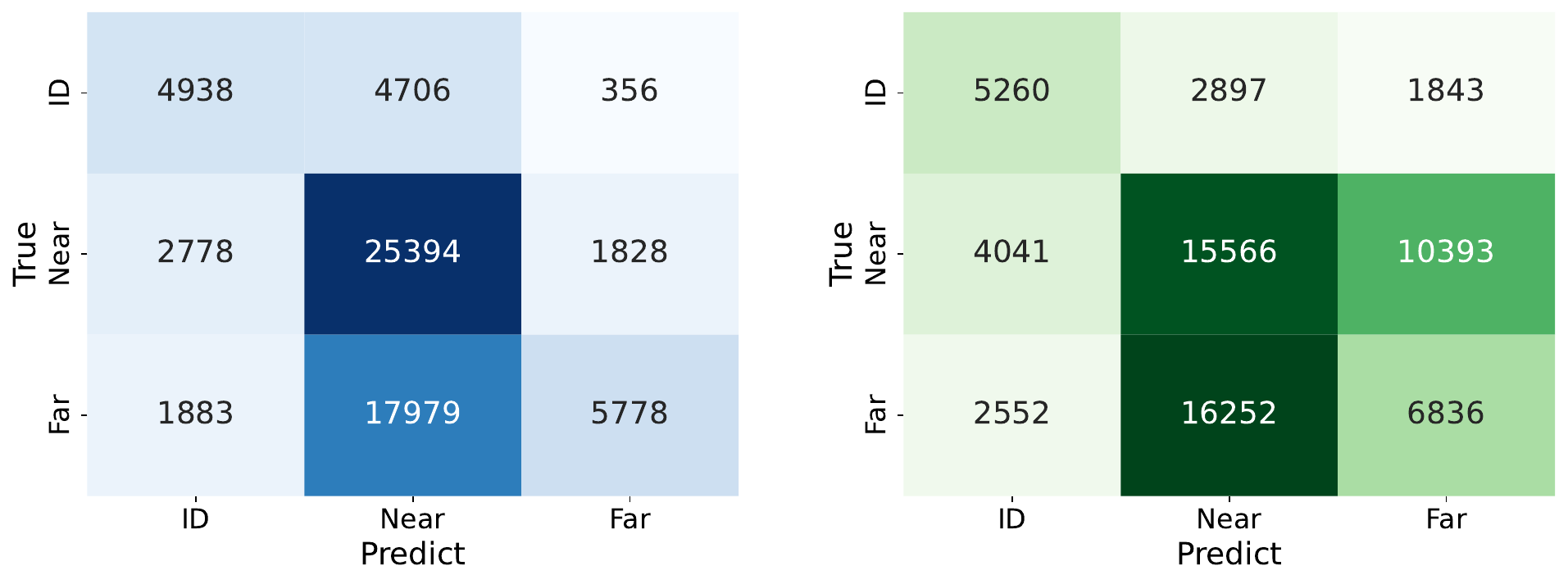} 
    \caption{Comparison of confusion matrices for our method (left) and PALM (right). Our method shows a cleaner block-diagonal structure and significantly reduces high-risk misclassifications (e.g., Near-OOD predicted as ID).}
    \label{fig:confusion_matrix}
\end{figure}

\paragraph{Analysis of Main Results} 
The results in Table~\ref{tab:main_results_final} confirm our framework's superiority from multiple angles. First, Ours SSV achieves sota performance among all deployable methods, securing the highest Macro F1 (0.53) and a standout Near-OOD F1 (0.64), which directly mitigates the cognitive flattening problem. Second, the decisive role of manifold quality is validated by comparing Ours with PALM; our hierarchically-aware backbone boosts the nSR from 0.4265 to a far superior 0.3268, proving that a well-structured representation is the primary driver of performance. This superiority is visually corroborated by Figure~\ref{fig:confusion_matrix}, which shows a dramatic reduction in high-risk Near-OOD to ID misclassifications from 4,041 to 2,778.

\subsection{Ablation Studies}

\paragraph{Ablation on Loss Component.}
First, we validate the cornerstone of our framework $\mathcal{L}_{\text{Hierarchy}}$, by comparing our full model against a variant trained without it. The results in Table~\ref{tab:ablation_loss_extended} are unequivocal. Removing the hierarchical loss triggers a geometric collapse: the manifold's Cohesion score deteriorates dramatically (from 0.8650 to 1.1670), regressing to the semantic collapse flaw. This geometric degradation translates directly to a catastrophic failure in the downstream task, with the nSR score plummeting from 0.3268 to 0.4265. This confirms that $\mathcal{L}_{\text{Hierarchy}}$ is the fundamental mechanism that engineers a superior manifold and enables fine-grained risk stratification.

\begin{table}[htbp]
    \centering
    \setlength{\tabcolsep}{4pt}
    \renewcommand{\arraystretch}{1.2}
    \begin{tabular}{l|ccc|cc}
        \toprule
        \textbf{Model} & \textbf{Com} $\downarrow$ & \textbf{Coh} $\downarrow$ & \textbf{Sep} $\uparrow$ & \textbf{nSR} $\downarrow$ & \textbf{F1} $\uparrow$ \\
        \midrule
        w/o $\mathcal{L}_{\text{Hierarchy}}$ & 0.4536 & 1.1670 & \textbf{1.1684} & 0.4265 & 0.4222 \\
        \textbf{Ours} & \textbf{0.3595} & \textbf{0.8650} & 
        0.8632 & \textbf{0.3268} & \textbf{0.5304} \\
        \bottomrule
    \end{tabular}
    \caption{Ablation study on the hierarchical loss component. Removing the loss leads to a catastrophic degradation in both manifold quality (especially Cohesion) and downstream task performance.}
    \label{tab:ablation_loss_extended}
\end{table}

\paragraph{Ablation Study on SSV's Discriminative Power.}
\label{sec:ssv_discriminative_power}

Second, we dissect the contribution of each SSV component and its dependence on manifold quality in Table~\ref{tab:ablation_ssv_combined}. On our hierarchically-structured manifold, the SSV components are clearly synergistic, with their combination in the Full SSV culminating in the optimal performance (nSR: 0.3268). In stark contrast, this synergy vanishes on PALM's geometrically flat manifold, where simpler feature combinations outperform the Full SSV. This is powerful evidence that on a flawed manifold, adding more diagnostic dimensions introduces noise rather than clarity, reinforcing our central thesis: a meticulously shaped semantic manifold is a prerequisite for unlocking the full diagnostic power of a multi-dimensional probe like SSV.

\begin{table}[htbp]
    \centering
    \setlength{\tabcolsep}{4pt} 
    \renewcommand{\arraystretch}{1} 
    \small 
    \begin{tabular}{l|cc|cc}
        \toprule
        \multirow{2}{*}{\textbf{SSV Components}} & \multicolumn{2}{c|}{\textbf{Ours-Backbone}} & \multicolumn{2}{c}{\textbf{PALM-Backbone}} \\
        \cmidrule(lr){2-3} \cmidrule(lr){4-5}
        & nSR$\downarrow$ & Macro F1$\uparrow$ & nSR$\downarrow$ & Macro F1$\uparrow$ \\
        \midrule
        $S_{\text{conf}}$  & 0.3695 & 0.4822 & 0.4083 & 0.3981 \\
        $S_{\text{novel}}$  & 0.3298 & 0.3885 & 0.4287 & 0.3455 \\
        $S_{\text{ambig}}$  & 0.3598 & 0.4078 & 0.5015 & 0.3534 \\
        \midrule
        $S_{\text{conf}} + S_{\text{novel}}$ & 0.3273 & 0.5300 & 0.4499 & 0.4072 \\
        $S_{\text{conf}} + S_{\text{ambig}}$ & 0.3483 & 0.5050 & 0.4154 & 0.4029 \\
        $S_{\text{novel}} + S_{\text{ambig}}$ & 0.3365 & 0.4639 & 0.4206 & 0.4240 \\
        \midrule
        \textbf{Full SSV} & \textbf{0.3268} & \textbf{0.5304} & \textbf{0.4265} & \textbf{0.4222} \\
        \bottomrule
    \end{tabular}
    \caption{SSV Dimension Ablation Study. The table compares the performance of SSV component combinations on our backbone versus the PALM backbone.}
    \label{tab:ablation_ssv_combined}
\end{table}

\subsection{Generalization and Broader Impact}
\label{sec:generalization_broader_impact}

\subsubsection{Robustness Across Diverse OOD Scenarios}
\label{sec:exp_robustness}
To rigorously assess the robustness and generalizability of our framework, we conducted a large-scale sensitivity analysis, evaluating our method against the strong PALM baseline across a matrix of 20 challenging Near-OOD and Far-OOD dataset combinations. 
The comprehensive results are detailed in Table~\ref{tab:sensitivity_analysis_side_by_side} and visualized in Appendix. The findings unequivocally demonstrate the superiority of our approach. Our method achieves a lower (better) nSR in \textbf{17 out of 20 (85\%)} of the tested configurations, often by a significant margin. For the sake of transparent analysis, we acknowledge that in 3 cases, primarily when Places365 is the Far-OOD dataset, the PALM baseline shows a marginal advantage, as can be seen in the table. However, these few instances are outweighed by the overwhelming consistency of our method's superiority, with an average performance delta of a robust \textbf{+0.058} across all 20 experiments. This comprehensive study validates that the benefits derived from our hierarchically-aware manifold constitute a fundamental and more generalizable solution for semantic risk stratification.

\begin{table}[htbp]
    \centering
    \small 
    \renewcommand{\arraystretch}{1} 
    \setlength{\tabcolsep}{2pt} 
     \begin{tabular}{l c c @{\hspace{3pt}} c c @{\hspace{3pt}} c c @{\hspace{3pt}} c c}
        \toprule
        \multirow{2}{*}{\diagbox[width=4.5em, trim=l]{\textbf{Far}}{\textbf{Near}}} & \multicolumn{2}{c}{\textbf{LSUN-F}} & \multicolumn{2}{c}{\textbf{ImageNet-F}} & \multicolumn{2}{c}{\textbf{ImageNet-R}} & \multicolumn{2}{c}{\textbf{CIFAR-10}} \\
        \cmidrule(lr){2-3} \cmidrule(lr){4-5} \cmidrule(lr){6-7} \cmidrule(lr){8-9}
        & PALM & Ours & PALM & Ours & PALM & Ours & PALM & Ours \\
        \midrule
        \textbf{Places365} & 0.47 & \textbf{0.40} & \textbf{0.46} & 0.53 & \textbf{0.50} & 0.52 & 0.53 & \textbf{0.51} \\
        \textbf{LSUN}      & 0.55 & \textbf{0.47} & 0.57 & \textbf{0.47} & 0.62 & \textbf{0.52} & 0.53 & \textbf{0.47} \\
        \textbf{Texture}       & 0.46 & \textbf{0.34} & 0.47 & \textbf{0.40} & 0.52 & \textbf{0.41} & 0.51 & \textbf{0.39} \\
        \textbf{iSUN}      & 0.52 & \textbf{0.42} & 0.51 & \textbf{0.49} & 0.59 & \textbf{0.56} & \textbf{0.51} & 0.52 \\
        \textbf{SVHN}      & 0.33 & \textbf{0.29} & 0.42 & \textbf{0.29} & 0.45 & \textbf{0.30} & 0.44 & \textbf{0.36} \\
        \bottomrule
    \end{tabular}
    \caption{OOD Dataset Sensitivity Analysis: nSR performance of our method versus PALM across various Near-OOD and Far-OOD combinations. All values are rounded to two decimal places.}
    \label{tab:sensitivity_analysis_side_by_side}
\end{table}

\begin{table}[htbp]
    \centering
    
    \small
    \renewcommand{\arraystretch}{1.0}
    \setlength{\tabcolsep}{3pt}
    \begin{tabular}{l|cc|cc||cc}
        \toprule
        \multirow{2}{*}{\textbf{Method}} & \multicolumn{2}{c|}{\textbf{SVHN}} & \multicolumn{2}{c||}{\textbf{LSUN}} & \multicolumn{2}{c}{\textbf{Average}} \\
        & FPR$\downarrow$ & AUROC$\uparrow$ & FPR$\downarrow$ & AUROC$\uparrow$ & FPR$\downarrow$ & AUROC$\uparrow$ \\
        \midrule
        MSP & 78.89 & 79.80 & 83.47 & 75.28 & 81.18 & 77.54 \\
        VIM & 73.42 & 84.62 & 86.96 & 69.74 & 80.19 & 77.18 \\
        ODIN & 70.16 & 84.88 & 76.36 & 80.10 & 73.26 & 82.49 \\
        Energy & 66.91 & 85.25 & 59.77 & 86.69 & 63.34 & 85.97 \\
        VOS & 43.24 & 82.80 & 73.61 & 84.69 & 58.43 & 83.75 \\
        CSI & 44.53 & 92.65 & 75.58 & 83.78 & 60.06 & 88.22 \\
        SSD+ & 31.19 & 94.19 & 79.39 & 85.18 & 55.29 & 89.69 \\
        kNN+ & 39.23 & 92.78 & 48.99 & 89.30 & 44.11 & 91.04 \\
        CIDER & 12.55 & 97.83 & 30.24 & 92.79 & 21.40 & 95.31 \\
        NPOS & 10.62 & 97.49 & 20.61 & 92.61 & 15.62 & 95.05 \\
        PALM & \underline{3.03} & \underline{99.23} & \underline{10.58} & \underline{97.70} & \underline{6.81} & \underline{98.47} \\
        \textbf{Ours} & \textbf{2.20} & \textbf{99.56} & \textbf{4.15} & \textbf{98.94} & \textbf{3.18} & \textbf{99.25} \\
        \bottomrule
    \end{tabular}
    \caption{Performance on key binary OOD benchmarks (SVHN, LSUN). Our method achieves the best average performance and sota results on both individual datasets. Full results are in Appendix.}
    \label{tab:binary_ood_summary}
\end{table}

\subsubsection{Performance on Conventional Binary OOD Detection}
\label{sec:binary_ood_performance}
To validate the generalizability of our learned representations, we benchmarked our framework on the conventional binary OOD detection task. While the full comparison against a suite of strong baselines across four common OOD datasets is detailed in Appendix, we highlight key findings here.
The results in Table~\ref{tab:binary_ood_summary} reveal a nuanced and compelling story. First, on several key benchmarks, our method achieves sota results, outperforming highly specialized methods. Notably, on SVHN and LSUN, our model attains the best performance across both FPR95 (at \textbf{2.20\%} and \textbf{4.15\%}, respectively) and AUROC (at \textbf{99.56\%} and \textbf{98.94\%}). This showcases the powerful discriminative capability of the representations learned through our hierarchical objective.
Second, the competitive average performance seen in the full results table should be interpreted in the context of our framework's design. Our model is optimized for a significantly more complex, hierarchical task. Therefore, its strong performance on this simpler binary task—without any specific tuning—is not a limitation but a testament to the inherent robustness and versatility of the learned feature space. This demonstrates that by solving a harder, more structured problem, our method learns a fundamentally sound representation that avoids narrow overfitting to a single objective.

\section{Conclusion}
\label{sec:conclusion}
This paper addresses the cognitive flattening limitation in conventional OOD detection by reframing the problem as a principled framework that quantifies Semantic Surprise. We introduce Low-Entropy Semantic Manifolds as the ideal knowledge structure and a Hierarchical Prototypical Network to construct them, developing the SSV for multi-dimensional, interpretable risk diagnosis. Our experiments validated this approach, showing that geometric manifold quality is the critical driver for fine-grained OOD detection. The SSV framework, with its multi-faceted diagnostic report, fundamentally surpasses the theoretical performance ceiling of single-score methods. The intentional push-pull dynamic in our loss function leads to a robust and generalizable representation, and our proposed nSR provides a more rational standard for evaluation.
Despite relying on a pre-defined class hierarchy, the SSV's diagnostic capability is highly transferable. Future research could explore automatically learning these semantic structures, possibly with hyperbolic geometry or knowledge graphs, and applying SSV to domains like continual learning and active learning. By reframing the OOD problem, this work offers a robust and interpretable solution for AI safety, opening a new, cognitively-aligned research avenue.

\bibliography{aaai2026}

\end{document}